\documentclass[10pt,twocolumn,letterpaper]{article}

\usepackage{cvpr}              % To produce the CAMERA-READY version
\usepackage[dvipsnames,svgnames,table]{xcolor}

\definecolor{cvprblue}{rgb}{0.21,0.49,0.74}
\usepackage[pagebackref,breaklinks,colorlinks,citecolor=cvprblue]{hyperref}
\usepackage{amssymb}
\usepackage{multirow}
\usepackage{circledsteps}

\def\confName{CVPR}
\def\confYear{2024}

\ifdefined \GramaCheck
  \newcommand{\CheckRmv}[1]{}
  \renewcommand{\eqref}[1]{Eq. (1)}
  \renewcommand{\equref}[1]{Eq. (1)}
  \renewcommand{\figref}[1]{Figure 1}
  \renewcommand{\tabref}[1]{Table 1}
\else
  \newcommand{\CheckRmv}[1]{#1}
  \renewcommand{\eqref}[1]{Eq.~(\ref{#1})}
\fi
\graphicspath{{./figures/}}
\DeclareGraphicsExtensions{.pdf,.jpg,.png,.bmp}

\definecolor{rgb1}{RGB}{214,  38, 40}   % 1 ceiling
\definecolor{rgb2}{RGB}{43, 160, 4}     % 2 floor
\definecolor{rgb3}{RGB}{158, 216, 229}  % 3 wall
\definecolor{rgb4}{RGB}{114, 158, 206}  % 4 window
\definecolor{rgb5}{RGB}{204, 204, 91}   % 5 chair
\definecolor{rgb6}{RGB}{255, 186, 119}  % 6 bed
\definecolor{rgb7}{RGB}{147, 102, 188}  % 7 sofa
\definecolor{rgb8}{RGB}{30, 119, 181}   % 8 table
\definecolor{rgb9}{RGB}{160, 188, 33}   % 9 tvs mod
\definecolor{rgb10}{RGB}{255, 127, 12}  % 10 furn
\definecolor{rgb11}{RGB}{196, 175, 214} % 11 objects
\usepackage{pifont}
\newcommand{\cmark}{\ding{51}}%
\newcommand{\xmark}{\ding{55}}%

\title{Unleashing Network Potentials for Semantic Scene Completion}

\author{Fengyun Wang$^1$ \ Qianru Sun$^2$ \ Dong Zhang$^3$ \ Jinhui Tang$^1$\thanks{Corresponding author.} \\
{\small $^1$School of Computer Science and Engineering, Nanjing University of Science \& Technology}\\
{\small $^2$Singapore Management University \ $^3$The Hong Kong University of Science \& Technology}\\ 
{\small E-mail: \{fereenwong, jinhuitang\}@njust.edu.cn; qianrusun@smu.edu.sg; dongz@ust.hk}}
\begin{document}
\maketitle
\begin{abstract}
Semantic scene completion (SSC) aims to predict complete 3D voxel occupancy and semantics from a single-view RGB-D image, and recent SSC methods commonly adopt multi-modal inputs. However, our investigation reveals two limitations: ineffective feature learning from single modalities and overfitting to limited datasets. To address these issues, this paper proposes a novel SSC framework - Adversarial Modality Modulation Network (AMMNet) - with a fresh perspective of optimizing gradient updates. The proposed AMMNet introduces two core modules: a cross-modal modulation enabling the interdependence of gradient flows between modalities, and a customized adversarial training scheme leveraging dynamic gradient competition.
Specifically, the cross-modal modulation adaptively re-calibrates the features to better excite representation potentials from each single modality. The adversarial training employs a minimax game of evolving gradients, with customized guidance to strengthen the generator's perception of visual fidelity from both geometric completeness and semantic correctness. Extensive experimental results demonstrate that AMMNet outperforms state-of-the-art SSC methods by a large margin, providing a promising direction for improving the effectiveness and generalization of SSC methods. \underline{Our code is available at this \href{https://github.com/fereenwong/AMMNet}{link}}.
\end{abstract}
    
\section{Introduction}
\label{sec:intro}
\begin{figure}[!t]
\centering
\includegraphics[width=0.48\textwidth]{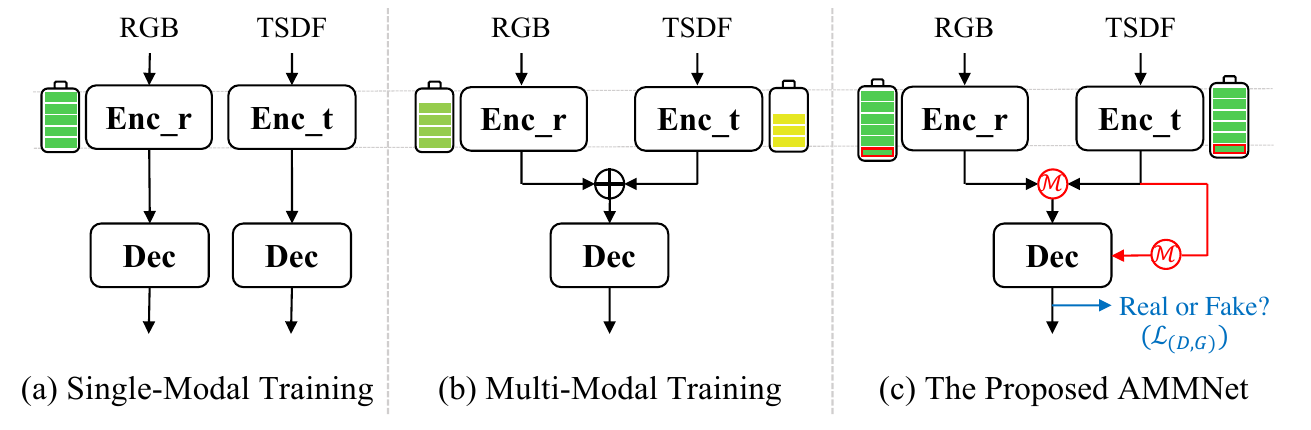}
\caption{
\textbf{Comparisons of encoder representation power}. For the multi-modal training in (b), the informative representations in RGB and TSDF are not fully unleashed compared to learning them individually in (a). The proposed Adversarial Modality Modulation Network (AMMNet) in (c) enables a more thorough unleashing of potentials via \textcolor{red}{``cross-modal modulation $\mathcal{M}$''} and \textcolor[RGB]{53,114,188}{``adversarial training $\mathcal{L}_{(D,G)}$''}.} 
\label{fig:teaser}
\vspace{-3mm}
\end{figure}

% -------------------------------
Semantic scene completion (SSC) is a crucial task in 3D scene understanding domain that seeks to forecast complete 3D voxel occupancy and semantics from a single-view RGB-D image~\cite{liu2018see,li2019rgbd,liu20203d}.
Current SSC methods rely on multi-modal inputs like RGB images and depth represented as Truncated Signed Distance Function (TSDF) representations, as shown in Figure ~\ref{fig:teaser} (b), which provide complementary cues for scene geometry reconstruction and semantic prediction. These methods typically utilize an encoder-decoder paradigm to exploit multi-modal data, where the RGB image and TSDF are encoded separately and subsequently combined for final predictions. Despite the promising results demonstrated by these multi-modal models on indoor scene completion benchmarks~\cite{silberman2012indoor}, two key observations can still be made from these models~\cite{chen20203d,wang2023semantic}.

% ---------------------------------
% ---------------------------------

% ---------------------------------
\CheckRmv{
\begin{figure*}[!t]
\centering
\includegraphics[width=0.98\textwidth]{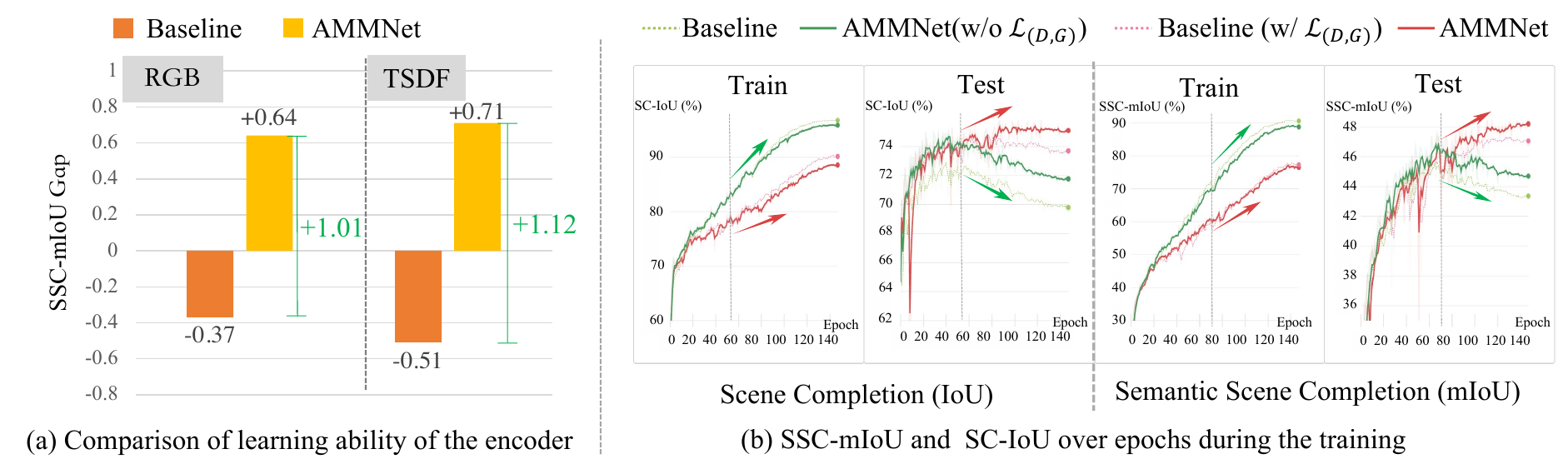}
\caption{\textbf{Two key observations on multi-modal SSC models}. \textbf{(a)} Performance drops of \textcolor[RGB]{219,129,67}{``multi-modal encoders''} compared to single-modal counterparts, validating insufficient unleashing of modalities in joint training. \textcolor[RGB]{245,194,66}{``Our method''} demonstrates significantly enhanced encoder capabilities. \textbf{(b)} \textcolor[RGB]{79,173,91}{``Diverging training/validation curves''} of baseline and the variant AMMNet, indicating overfitting issues. Under the adversarial training scheme $\mathcal{L}_{(D,G)}$, our model alleviates overfitting and achieves \textcolor{red}{``steadily increasing performance''}.}
\label{fig:problem}
\end{figure*}
}
% ---------------------------------

\noindent \textbf{Observation 1:} \emph{When learned jointly in multi-modal models, the rich information in individual modalities is not sufficiently unleashed compared to single-modal models.}
To validate this observation, we conducted a two-stage experiment. In the first stage, we trained uni-modal and multi-modal SSC models using the same dataset and settings. In the second stage, we evaluated separate single-modal networks initialized with encoder weights from stage one models to assess the learned representations.
Through controlled comparison, we found degraded representation capabilities of the multi-modal encoders compared to their uni-modal counterparts. Specifically, we examined the representation power of RGB and TSDF encoders by initializing two RGB-only networks and two TSDF-only networks using corresponding encoder weights from uni-modal and multi-modal stage one models. With the encoders frozen, the decoders were then trained. As the only difference was the frozen encoder, performance gaps demonstrated the insufficient unleashing of modalities in multi-modal training.
As shown in Figure~\ref{fig:problem} (a), employing the multi-modal RGB encoder led to a performance drop of $0.37\%$ in terms of SSC-mIoU compared to utilizing the single-modal RGB encoder, on the NYU dataset~\cite{silberman2012indoor}. Similarly, adopting the multi-modal TSDF encoder incurred a $0.51\%$ decrease in SSC-mIoU compared to the single-modal TSDF encoder.

\noindent \textbf{Observation 2:} \emph{Deep SSC models trained with limited scene data are prone to overfitting.}
To validate this observation, we examined the training processes of a baseline model~\cite{wang2023semantic} and a variant of our AMMNet without the proposed adversarial training scheme $\mathcal{L}_{(D,G)}$. We observed severe overfitting behaviors, where models first reached optimal validation performance but further training led to increasing divergence between training and validation.
As illustrated in Figure~\ref{fig:problem}, the baseline model~\cite{wang2023semantic} (dotted green line) achieved the best validation score in the middle of training, while later epochs led to a 22.8\% SSC-mIoU increase on the training set but a 2.8\% SSC-mIoU drop on the validation set. A similar divergence was observed for the AMMNet \emph{w/o} $\mathcal{L}_{(D,G)}$ (solid green line), validating that SSC models tend to overfit the training data.

To address these issues, we propose an Adversarial Modality Modulation Network (AMMNet), a novel SSC framework to better unleash the potential by optimizing gradient updating. A conceptual illustration is presented in Figure \ref{fig:teaser} (c), it consists of two key components designed to address the identified issues.
1)~A cross-modal modulation is introduced to better unleash the potentials of individual modalities. With inter-dependent gradient updating across modalities, it can stimulate the encoders to fully unleash the representations of RGB and TSDF in joint training. Specifically, it adaptively recalibrates the RGB features by incorporating information from the TSDF. As illustrated in Figure~\ref{fig:problem}~(a), compared to the encoder in the multi-modal baseline~\cite{wang2023semantic} on the NYU~\cite{silberman2012indoor} dataset, the RGB and TSDF encoders in AMMNet demonstrate improved capabilities, with $1.01\%$ and $1.12\%$ higher SSC-mIoU, respectively. 
2)~A customized adversarial training scheme is developed to alleviate overfitting. The minimax competition in the scheme dynamically stimulates the continuous evolution of the models. To provide effective supervision, particularly for SSC, we construct two types of perturbed ground truths: one with disrupted geometric completeness and the other with randomly shuffled semantic categories. These perturbed ground truths are fed to the discriminator as fake samples, explicitly enhancing the discriminator's ability to recognize flaws in both geometry and semantics. Figure~\ref{fig:problem}~(b) shows that by incorporating the proposed adversarial training scheme, both the baseline model~\cite{wang2023semantic} and our AMMNet achieve steadily increasing training and validation accuracy over epochs. 

Our main contributions of AMMNet are thus two-fold.
1) we proposed a cross-modal modulation module to better exploit single-modal representations in multi-modal learning.
2) we developed a customized adversarial training scheme to prevent overfitting. Experimental results validated the effectiveness of our AMMNet, showing that it outperforms state-of-the-art SSC methods by significant margins, \eg, improving SSC-mIoU by 3.5\% on NYU~\cite{silberman2012indoor} and 3.3\% on NYUCAD~\cite{firman2016NYUCAD} compared to previous methods.
% 

% -----------------------------------
\CheckRmv{
\begin{figure*}[!t]
\centering
\includegraphics[width=0.99\textwidth]{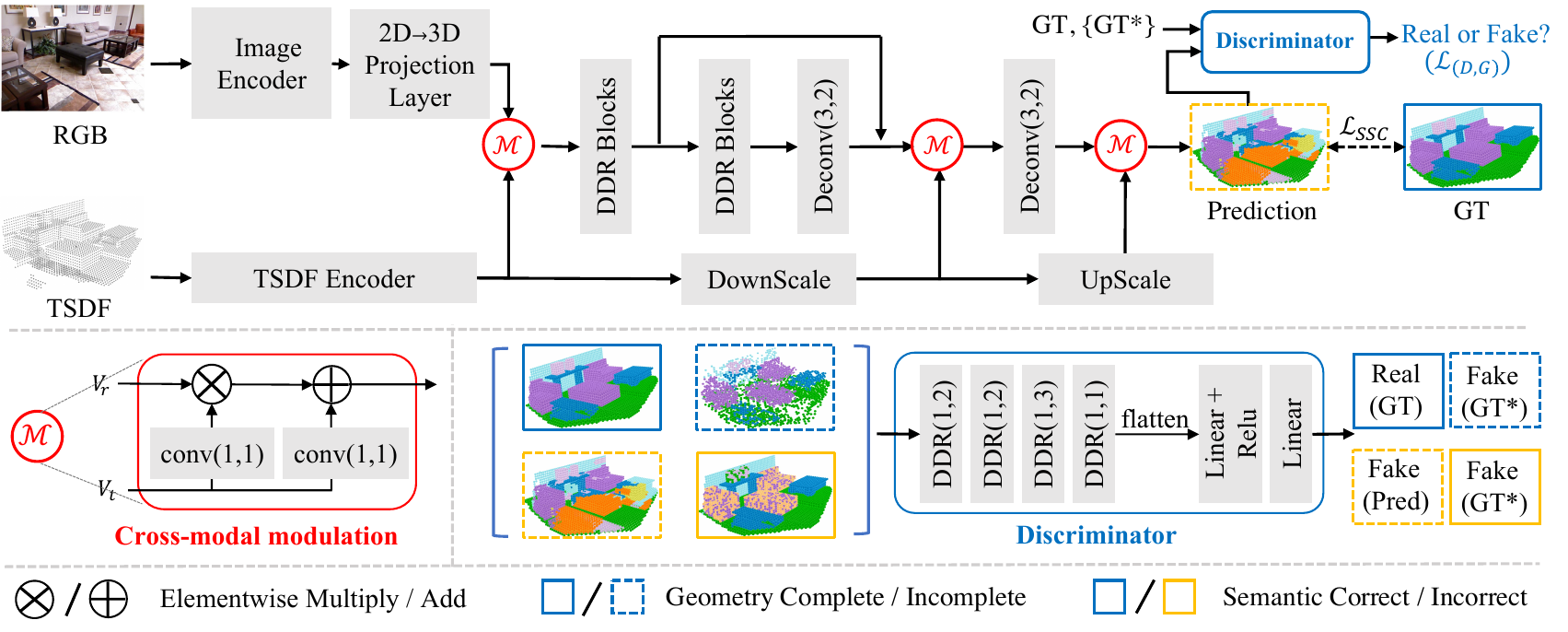}
\caption{\textbf{The overall framework of our AMMNet}. It consists of three components: an image encoder for RGB input, a TSDF encoder for TSDF input, and a decoder for final prediction. It has two novel modules: cross-modal modulations after the encoders and decoder to recalibrate features, and a discriminator that distinguishes real/fake voxels to mitigate overfitting issues.
The conv($k$, $s$)/Deconv($k$, $s$) denotes 3D conv/deconv layer with kernel size $k$ and stride $s$, and DDR($d$, $s$) denotes DDR layer~\cite{li2019rgbd} with dilation $d$ and stride $s$.}
\label{fig:framework}
\end{figure*}
}
% -----------------------------------

% 
% 
\section{Related Work}
\label{sec:relatedwork}
% ----------------------------------
\noindent
\textbf{Semantic Scene Completion (SSC).}
SSC can be roughly categorized into single-modal methods and multi-modal methods based on the usage of input modalities.
Single-modal methods take only TSDF~\cite{song2017semantic,dourado2019edgenet} or points~\cite{zhong2020semantic} converted from depth as input. Volume-based methods like SSCNet~\cite{song2017semantic} adopt 3D CNNs on TSDF. Point-based methods like SPCNet~\cite{zhong2020semantic} avoid voxel discretization but are prone to noise. Single-modal inputs are limited in complementary cues.
Multi-modal methods can be further grouped. Some methods process depth as 2D images using 2D CNNs~\cite{guo2018_VVNet,liu2018see,li2019rgbd,li2020anisotropic}, which are less effective for 3D geometry.
Other works encode depth as points~\cite{rist2021semantic,tang2022not} or TSDF~\cite{chen20203d,wang2023semantic}, and adopt dual-branch networks fusing RGB, TSDF, and/or points, achieving state-of-the-art performance.
However, they overlook the problems of insufficient encoder feature learning and overfitting as observed in our study. Our work is the first tailored solution addressing these limitations for advanced multi-modal SSC.

\noindent
\textbf{Multi-Modal Learning.}
Multi-modal learning has attracted increasing attention owing to the growing availability of multi-modal data. However, it does not always achieve synergistic performance surpassing the sum of individual modalities~\cite{peng2022balanced}. The crux lies in the inadequate harnessing of all modal information by most joint training methods~\cite{ramachandram2017deep}. 
Prior works have explored various directions to address this challenge. Hu et al. proposed a temporal multi-modal deep learning architecture that transforms connected multi-modal restricted Boltzmann machines into probabilistic sequence models, achieving more efficient joint feature learning~\cite{hu2016temporal}.
Du et al. referred to inferior joint training outcomes as "modality failure" and proposed combining fusion objectives with uni-modal distillation~\cite{du2021improving}.
Instead of designing complex joint training architectures, we aim to enable fuller unleashing of modal representations by proposing simple yet effective solutions - cross-modal modulation and adversarial training.

\noindent
\textbf{Overfitting Problem in Deep Learning.}
Various strategies have been explored to alleviate the overfitting problem in deep neural networks. They can be grouped into three categories~\cite{bejani2021systematic}:
1)~\emph{Passive Methods}, like neural architecture search~\cite{zoph2016neural} and ensembling~\cite{ganaie2022ensemble}, aim to determine suitable model configurations before training starts. However, such static configurations lack the flexibility to adapt to dynamic training processes.
2)~\emph{Active Methods}, include techniques like Dropout~\cite{srivastava2014dropout}, augmentation~\cite{shorten2021text}, and Normalization~\cite{garbin2020dropout}. In this category, the model architecture remains unchanged, but certain model components are unleashed during each training step. 
3)~\emph{Semi-active Methods}, like pruning~\cite{luo2017thinet} and network construction~\cite{du2010fast}, updates the network by adding or removing neurons or connections. Utilizing an adversarial training scheme also falls under this category. Unlike other methods requiring complex network search or design, our adversarial training scheme uses a minimax game to stimulate the model evolution, alleviating overfitting without architectural optimization. To our best knowledge, this is the first attempt at utilizing an adversarial training scheme to mitigate overfitting in SSC.

% --------------------------------
\section{Preliminaries}
% --------------------------------
The baseline model~\cite{wang2023semantic} for SSC typically contains an image encoder for RGB input, a TSDF encoder for TSDF input, and a decoder for final prediction. 
Specifically, the RGB encoder extracts a $D$-channel 2D feature $\mathbf{F}_r$ from the RGB input. To obtain 3D representations, a 2D-3D projection layer is applied to map $\mathbf{F}_r$ to visible surfaces based on per-pixel depth values, while filling other areas with zeros. This results in a 3D RGB feature $\mathbf{V}_r \in \mathbb{R}^{D \times G_x \times G_y \times G_z}$, where $G_x$, $G_y$, and $G_z$ denote the width, height, and depth of the voxel grid.
The TSDF encoder employs 3D convolutions to encode the input voxels to 3D feature maps $\mathbf{V}_t \in \mathbb{R}^{D \times G_x \times G_y \times G_z}$.
The projected 3D RGB feature $\mathbf{V}_r$ is fused with $\mathbf{V}_t$ via element-wise addition as: 
\begin{equation}
\bar{\mathbf{V}}_{cm} = \mathbf{V}_r + \mathbf{V}_t.
\label{eq:featfusion}
\end{equation}
The fused feature $\bar{\mathbf{V}}_{cm}$ is fed into the decoder, which applies 3D convolutions with skip connection to generate the completed voxel grid $\hat{Y} \in \mathbb{R}^{(C + 1) \times G_x \times G_y \times G_z}$, where $C$ is the number of classes and $+1$ is an additional channel indicating voxel occupancy, \emph{i.e.}, empty or not.

This structure has been widely adopted as a basic framework for multi-modal SSC models, such as \cite{chen20203d,wang2023semantic}. Such a simple addition fails to fully unleash the potential of each modality as illustrated in observation 1.

% --------------------------------
\section{Adversarial Modality Modulation Network}
\label{sec:method}
% --------------------------------
In this section, we will elaborate on the proposed AMMNet. We introduce two key components in AMMNet to address the aforementioned limitations of existing methods: a cross-modal modulation module and a customized adversarial training scheme. 
% --------------------------------
\subsection{Cross-Modal Modulation}
% --------------------------------
The cross-modal modulation is expressed as a red circled $\mathcal{M}$ in Figure \ref{fig:framework}.
It adaptively recalibrates the RGB features by incorporating information from the TSDF, which enables an interdependent gradient updating across modalities. 
It first transforms $\mathbf{V}_t$ into a scale $\mathbf{M}_s$ and a bias $\mathbf{M}_b$ using two $1 \times 1 \times 1$ convolutional layers. $\mathbf{M}_s$ and $\mathbf{M}_b$ share the same shape as $\mathbf{V}_r$. Then, it modulates $\mathbf{V}_r$ via an element-wise operation, which is expressed as:
% --------------------------------
\begin{equation}
\hat{\mathbf{V}}_{cm} = \mathbf{V}_r \otimes (1 + \sigma(\mathbf{M}_s)) + \mathbf{M}_b.
\end{equation}
% --------------------------------
where $\otimes$ denotes element-wise multiplication, and $\sigma$ is the activation function $sigmoid$. The recalibrated feature $\hat{\mathbf{V}}_{cm}$ is then fed to the decoder. 
Compared to the element-wise addition operation in the baseline model~\cite{wang2023semantic}, our modulation enables interdependent gradient updating across modalities. 
Specifically, the gradient, w.r.t. $\mathbf{V}_r$, in the modulation is formulated as:
% --------------------------------
\begin{equation}
\frac{\partial \mathcal{L}}{\partial \mathbf{V}_r} = \frac{\partial \mathcal{L}}{\partial \hat{\mathbf{V}}_{cm}} \otimes (1 + \sigma(\mathbf{M}_s))
\end{equation}
% --------------------------------
where $\mathcal{L}$ is the loss function. The gradient w.r.t. $\mathbf{V}_r$ depends on $\mathbf{V}_t$ via the scale term $\mathbf{M}_s$. Similarly, the gradient $\mathbf{V}_t$ depends on RGB contexts. In contrast, in the baseline model~\cite{wang2023semantic}, the gradient $\mathbf{V}_r$ is expressed as:
\begin{equation}
\frac{\partial \mathcal{L}}{\partial \mathbf{V}_r} = \frac{\partial \mathcal{L}}{\partial \bar{\mathbf{V}}_{cm}},
\end{equation}
% --------------------------------
which is independent of $\mathbf{V}_t$ and is not desirable.

Meanwhile, the cross-modal modulation enables an improved forward fusion of the modalities. We will validate through ablation studies in Sec.~\ref{ssec:discussion} that the interdependent gradient updating plays a more critical role than the improved modality fusion.
As shown in Figure \ref{fig:framework}, we apply the cross-modal modulation on the last two layers of the decoder features similarly. To match the resolution with the decoder features, the TSDF feature $\mathbf{V}_t$ first goes through a DownScale module, resulting in downscaled TSDF features $\mathbf{V}_t' \in \mathbb{R}^{D\times \frac{G_x}{2}\times \frac{G_y}{2}\times \frac{G_z}{2}}$.
The resulting feature $\mathbf{V}_t'$ modulates the second last decoder layer output. After that, $\mathbf{V}_t'$ further goes through an UpScale module to restore the size and is used to modulate the last decoder layer output in the same way.

% --------------------------------
\subsection{Adversarial Training}
% --------------------------------
To prevent overfitting, a customized discriminator is designed to introduce dynamic adversarial gradients. By competing against the generator (i.e., the voxel predictor in this work) in a minimax game, the discriminator provides continuous and adaptive supervision besides the static losses.
In SSC tasks, complete spatial structures and correct semantic perceptions are vital. Accordingly, we customize the discriminator in three aspects:

\noindent 
\textbf{Geometry Completeness}. 
To make the discriminator more aware of geometry completeness, we construct additional fake samples by perturbing the geometry of ground truth voxels.
Specifically, we randomly erase some non-empty voxels in the ground truth maps to ``empty'', by:
\begin{equation}
\bar{\mathbf{y}}_i^G = \begin{cases}
    \textit{empty}, & r^G_i |_{r^G_i \sim U(0,1)} \leq p^G \\
    \mathbf{y}_i, & otherwise,
\end{cases}
\end{equation}
where $r^G_i$ is a random number that follows a uniform distribution $U(0,1)$, $p^G$ is the erasing probability, $\mathbf{y}_i \in \mathbf{Y}$ is a voxel in the ground truth, $\bar{\mathbf{y}}_i^G \in \bar{\mathbf{Y}}^G$ is the geometrically perturbed counterpart. 
This operation damages the integrity of the 3D structures randomly. Using these examples (denoted as ``fake'') in training, we force the discriminator to learn to pay more attention to the completeness and continuity of geometric shapes, such as to feedback to and guide the SSC voxel predictor to achieve high completeness.

% ----------------------------------
\begin{table*}[t]
\footnotesize
\renewcommand\arraystretch{1.3} 
\setlength{\tabcolsep}{5pt}{ 
\begin{center}
\begin{tabular}{r| c|c c c|c c c c c c c c c c c c} 
\hline
\multirow{2}{*}{Methods} & \multirow{2}{*}{Inputs} & \multicolumn{3}{c|}{Scene Completion(\%)} & \multicolumn{12}{c}{SSC-mIoU(\%)} \\ \cline{3-17}
 & & Prec. & Recall & IoU & \cellcolor{rgb1}ceil. & \cellcolor{rgb2}floor & \cellcolor{rgb3}wall & \cellcolor{rgb4}win. & \cellcolor{rgb5}chair & \cellcolor{rgb6}bed & \cellcolor{rgb7}sofa & \cellcolor{rgb8}table & \cellcolor{rgb9}TVs & \cellcolor{rgb10}furn. & \cellcolor{rgb11}objs. & avg. \\ 
%  & \multicolumn{3}{c|}{scene completion} & \multicolumn{12}{c}{semantic scene completion} \\ \hline
% Methods & prec. & recall & IoU & ceil. & floor & wall & win. & chair & bed & sofa & table & tvs & furn. & objs. & avg. \\ 
\hline
% (ICCV2017)
SSCNet~\cite{song2017semantic} & D & 57.0 & \textbf{94.5} & 55.1 & 15.1 & 94.7 & 24.4 &  0.0 & 12.6 & 32.1 & 35.0 & 13.0 &  7.8 & 27.1 & 10.1 & 24.7\\
% 
% EsscNet~\cite{zhang2018efficient}  & 71.9 & 71.9 & 56.2 & 17.5 & 75.4 & 25.8 &  6.7 &  15.3 & 53.8 &  42.4 & 11.2 &    0 & 33.4 & 11.8 & 26.7\\ 
%  
% ForkNet~\cite{wang2019forknet} & D & - & - & 63.4 & 36.2 & 93.8 & 29.2 & 18.9 & 17.7 & 61.6 & 52.9 & 23.3 & 19.5 & 45.4 & 20.0 & 37.1 \\
% 
CCPNet~\cite{zhang2019cascaded-ccpnet}  & D & 74.2  & 90.8 & 63.5 & 23.5 & \textbf{96.3} & 35.7 & 20.2 & 25.8 & 61.4 & {56.1} & 18.1 & 28.1 & 37.8 & 20.1 & 38.5\\ 
\hline
DDRNet~\cite{li2019rgbd}  & RGB+D & 71.5  & 80.8 & 61.0 & 21.1 & 92.2 & 33.5 & 6.8 & 14.8 & 48.3 & 42.3 & 13.2 & 13.9 & 35.3 & 13.2 & 30.4\\ 
AIC-Net~\cite{li2020anisotropic} & RGB+D  &62.4&91.8& 59.2& 23.2 & 90.8 & 32.3 & 14.8 & 18.2  & 51.1  & 44.8 & 15.2   & 22.4 & 38.3  & 15.7  & 33.3 \\
% 
% TS3D~\cite{Garbade2018_twoStream} & RGB+D & - & - & 60.0 & 9.7 & 93.4 & 25.5 & 21.0 & 17.4 & 55.9 & 49.2 & 17.0 & 27.5 & 39.4 & 19.3 & 34.1 \\
% 
% SPCNet~\cite{zhong2020semantic} & Point &  72.1 & 42.2  & 36.3 & 33.8 & 64.4  & 38.3  & 7.5   & 30.7  & 53.4 &  42.6  & 19.7  & 5.5  & 34.2  & 13.9  & 31.3 \\
% IPF-SPCNet~\cite{zhong2020semantic} & RGB+D &  70.5  & 46.7  & 39.0 & 32.7  & 66.0 & 41.2 & 17.2 & 34.7 & 55.3  & 47.0 & 21.7 & 12.5  & 38.4  & 19.2  & 35.1 \\
% 
3D-Sketch~\cite{chen20203d} & RGB+D & 85.0 & 81.6 & 71.3 & 43.1 & 93.6 & 40.5 & 24.3 & 30.0 & 57.1 & 49.3 & 29.2 & 14.3 & 42.5 & 28.6 & 41.1 \\ 
%  
% IMENet~\cite{li2021imenet} & RGB+D & 90.0 & 78.4 & 72.1 & 43.6 & 93.6 & 42.9 & 31.3 & 36.6 & 57.6 & 48.4 & 32.1 & 16.0 & 47.8 & \textbf{36.7} & 44.2 \\ 
% 
FFNet~\cite{wang2022ffnet} & RGB+D & 89.3 & 78.5 & 71.8 & 44.0 & 93.7 & 41.5 & 29.3 & 36.2 & 59.0 & 51.1 & 28.9 & 26.5 & 45.0 & 32.6 & 44.4 \\
CleanerS~\cite{wang2023semantic} & RGB+D  & 88.0	& 83.5	& 75.0	& 46.3	& 93.9	& 43.2	& {33.7}	& 38.5	& {62.2}	& 54.8	& 33.7	& 39.2	& 45.7	& 33.8	&  47.7 \\
PCANet~\cite{li2023front} & RGB+D  & 89.5 & 87.5	& \textbf{78.9}	& 44.3	& 94.5	& 50.1	& 30.7	& 41.8	& \textbf{68.5}	& 56.4	& 32.6	& 29.9	& 53.6	& 35.4	& 48.9 \\
SISNet$_{\textrm{DLabv3}}$~\cite{cai2021semantic} & RGB+D  & 92.1	& 83.8	& 78.2	& 54.7	& 93.8	& \textbf{53.2}	& 41.9	& 43.6	& 66.2	& 61.4	& 38.1	& 29.8	& 53.9	& 40.3	& 52.4 \\
CVSformer$_{\textrm{DLabv3}}$~\cite{dong2023cvsformer} & RGB+D  & 87.7	&	82.1 & 73.7	& 46.3	& 94.3	& 43.5	& \textbf{42.8}	& 46.2	& 67.7	& 66.0	& 39.2	& 43.2	& 53.9	& 35.0	& 52.6 \\
\hline
AMMNet & RGB+D & \textbf{90.5}	& 82.1	& {75.6}	& {46.7}	& 94.2	& {43.9}	& 30.6	& {39.1}	& 60.3	& 54.8	& {35.7}	& {44.4}	& {48.2}	& 35.3	& {48.5} \\
AMMNet$_{\textrm{DLabv3}}$ & RGB+D & 88.7 & 84.5	& {76.3}	& \textbf{49.2}	& 94.2	& {47.0}	& {41.7}	& \textbf{52.4}	& 68.1	& \textbf{66.4}	& \textbf{46.4}	& \textbf{52.4}	& \textbf{58.3}	& \textbf{41.1}	& \textbf{56.1} \\
\hline
\end{tabular}
\vspace{-2mm}
\caption{Result comparisons on the test set of NYU~\cite{silberman2012indoor}. Results with “$_{\textrm{DLabv3}}$” denote that these results are based on DeepLabv3~\cite{chen2017rethinking} as the backbone network for RGB image feature extraction. Bold numbers represent the best performance.} 
\vspace{-5mm}
\label{tab:sota}
\end{center}}
\end{table*}

% ----------------------------------

\noindent 
\textbf{Semantic Correctness}. 
To force the discriminator to be more aware of semantic correctness, we construct another set of fake samples by shuffling semantic categories in the ground truth. For a scene with $m$ non-empty category labels, we randomly select $n \leq m$ categories to introduce perturbation, where $n$ is also randomly determined. The perturbation for a particular category $c_j$ can be formed as:
\begin{equation}
\bar{\mathbf{y}}^S_{i| i \in \{\mathbf{y}_i = c_j\}} = \begin{cases}
    Map_{c_j \rightarrow c_k, j\neq k}(\mathbf{y}_i), & r^S_i |_{r^S_i \sim U(0,1)} \leq p^S_j \\
    \mathbf{y}_i, & otherwise,
\end{cases}
\end{equation}
where $r^S_i$ is a random number that follows a uniform distribution $U(0,1)$, $p^S_j$ is a random perturbation probability for category $c_j$, $Map(\cdot)$ is an operation that transforms the categorical labels by mapping the class $c_j$ to a different class $c_k$, with $j,k=1,2,\dots, C$. Note that the category indices $j,k$ start from 1 instead of 0, as $c_0$ denotes empty voxels that remain unchanged. By discerning fakes with shuffled semantics, the discriminator is compelled to develop a more robust understanding of inter-class contexts.
Its adversarial gradients can thus feedback to and guide the SSC voxel predictor towards the increase of semantic correctness.

\noindent 
\textbf{Structurally Lightweight}. 
We adopt the lightweight DDR layer as a basic building unit to avoid excessive overheads. As illustrated in Figure~\ref{fig:framework}, the discriminator network $D$ consists of a series of DDR layers with different strides, followed by a few linear layers. It outputs a confidence score indicating whether the input voxel is real or fake, providing an overall understanding of the global structural layout and semantic coherence.
The overall adversarial training objective is thus as follows,
\begin{align}
&\mathcal{L}_{(D,G)} = \min\limits_{G} \max\limits_{D} [\mathbb{E}_{\mathbf{Y}}[\log D(\mathbf{Y})] + \mathbb{E}_{\hat{\mathbf{Y}}}[\log (1-D(\hat{\mathbf{Y}}))]] \nonumber \\
& \quad \quad +\mathbb{E}_{\bar{\mathbf{Y}}^G}[\log (1-D(\bar{\mathbf{Y}}^G))]] + \mathbb{E}_{\bar{\mathbf{Y}}^S}[\log (1-D(\bar{\mathbf{Y}}^S))]]
\end{align}
where $\hat{\mathbf{Y}}$ is the output of $G$, \ie, the voxel predictor. 
It tries to: 1) maximize the probability of correctly distinguishing real/fake voxels by $D$; 2) minimize the probability of generated voxels by $G$ being classified as real by $D$.

\subsection{Overall Loss Function}

The full training objective of the proposed AMMNet contains two parts: the SSC loss $\mathcal{L}_{SSC}$ and the proposed adversarial loss $\mathcal{L}_{(D,G)}$:
\begin{equation}
\mathcal{L}_{all} = \mathcal{L}_{SSC} + \beta \mathcal{L}_{(D,G)},
\end{equation}
where $\beta$ is a coefficient balancing the two parts.
$\mathcal{L}_{SSC}$ provides supervision for predicting voxels. Following \cite{wang2023semantic}, it contains two terms:
\begin{equation}
\mathcal{L}_{SSC} = SCE(\hat{\mathbf{Y}}, \mathbf{Y}) + \lambda SCE(\hat{\mathbf{Y}}_{2D},\mathbf{Y}_{2D}),
\end{equation}
where $SCE$ denotes the smooth cross entropy loss~\cite{szegedy2016rethinking}. The first term is applied to the 3D voxel predictions. The second term provides 2D supervision $\mathbf{Y}_{2D}$ for the image semantic prediction $\hat{\mathbf{Y}}_{2D}$. Specifically, $\hat{\mathbf{Y}}_{2D}$ is obtained by applying an additional convolutional layer on 2D RGB features, and $\mathbf{Y}_{2D}$ is obtained by back-projecting $\mathbf{Y}$ to 2D space as described in \cite{wang2023semantic}. $\lambda$ is a coefficient balancing the two loss terms.

% -------------------------------------
\begin{table*}[t]
\footnotesize
\renewcommand\arraystretch{1.3} 
\setlength{\tabcolsep}{5pt}{ 
\begin{center}
\begin{tabular}{r|c|c c c|c c c c c c c c c c c c} 
\hline
\multirow{2}{*}{Methods} & \multirow{2}{*}{Inputs} & \multicolumn{3}{c|}{Scene Completion(\%)} & \multicolumn{12}{c}{Semantic Scene Completion(\%)} \\ \cline{3-17}
 &  & Prec. & Recall & IoU & \cellcolor{rgb1}ceil. & \cellcolor{rgb2}floor & \cellcolor{rgb3}wall & \cellcolor{rgb4}win. & \cellcolor{rgb5}chair & \cellcolor{rgb6}bed & \cellcolor{rgb7}sofa & \cellcolor{rgb8}table & \cellcolor{rgb9}TVs & \cellcolor{rgb10}furn. & \cellcolor{rgb11}objs. & avg. \\ 
%  & \multicolumn{3}{c|}{scene completion} & \multicolumn{12}{c}{semantic scene completion} \\ \hline
% Methods & prec. & recall & IoU & ceil. & floor & wall & win. & chair & bed & sofa & table & tvs & furn. & objs. & avg. \\ 
\hline
% (ICCV2017)
SSCNet~\cite{song2017semantic} & D & 75.4 & \textbf{96.3} & 73.2 & 32.5 & 92.6 & 40.2 & 8.9 & 33.9 & 57.0 & 59.5 & 28.3 & 8.1 & 44.8 & 25.1 & 40.0\\
% 
% EsscNet~\cite{zhang2018efficient}  & 71.9 & 71.9 & 56.2 & 17.5 & 75.4 & 25.8 &  6.7 &  15.3 & 53.8 &  42.4 & 11.2 &    0 & 33.4 & 11.8 & 26.7\\ 
%  
% ForkNet~\cite{wang2019forknet} 
% 
CCPNet~\cite{zhang2019cascaded-ccpnet} & D & 91.3 & 92.6 & 82.4 & 56.2 & 94.6 & 58.7 & 35.1 & 44.8 & 68.6 & 65.3 & 37.6 & 35.5 & 53.1 & 35.2 & 53.2\\ 
\hline
DDRNet~\cite{li2019rgbd} & RGB+D & 88.7 & 88.5 & 79.4 & 54.1 & 91.5 & 56.4 & 14.9 & 37.0 & 55.7 & 51.0 & 28.8 & 9.2 & 44.1 & 27.8 & 42.8 \\ 
AIC-Net~\cite{li2020anisotropic} & RGB+D  & 88.2 & 90.3 & 80.5 & 53.0 & 91.2 & 57.2 & 20.2 & 44.6 & 58.4 & 56.2 & 36.2 & 9.7 & 47.1 & 30.4 & 45.8 \\
% 
% TS3D~\cite{Garbade2018_twoStream} & RGB+D & - & - & 76.1 & 25.9 & 93.8 & 48.9 & 33.4 & 31.2 & 66.1 & 56.4 & 31.6 & 38.5 & 51.4 & 30.8 & 46.2 \\
% 
% IPF-SPCNet~\cite{zhong2020semantic} & RGB+D & 83.3 & 72.7 & 63.5 & 58.8 & 91.9 & 60.5 & 25.2 & 53.6 & 72.9 & 62.4 & 33.8 & 12.4 & 53.6 & 32.5 & 50.7 \\
% 
3D-Sketch~\cite{chen20203d} & RGB+D & 90.6 & 92.2 & {84.2} & 59.7 & 94.3 & 64.3 & 32.6 & 51.7 & 72.0 & 68.7 & 45.9 & 19.0 & 60.5 & 38.5 & 55.2 \\ 
%  
% IMENet~\cite{li2021imenet} & RGB+D & 84.8 & 92.3 & 79.1 & - & - & - & - & - & - & - & - & - & - & - & 47.5 \\ 
% 
PCANet~\cite{li2023front} & RGB+D  & 92.1	&	91.8 & 84.3	& 54.8	& 93.1	& 62.8	& 44.3	& 52.3	& 75.6	& 70.2	& 46.9	& 44.8	& 65.3	& 45.8	& 59.6 \\
SISNet$_{\textrm{DLabv3}}$~\cite{cai2021semantic} & RGB+D  & 94.1	&	91.2 & \textbf{86.3}	& 63.4	& 94.4	& \textbf{67.2}	& 52.4	& 59.2	& 77.9	& 71.1	& 51.8	& 46.2	& 65.8	& \textbf{48.8}	& 63.5 \\
CVSformer$_{\textrm{DLabv3}}$~\cite{dong2023cvsformer} & RGB+D  & 94.0	&	91.0 & 86.0	& \textbf{65.6}	& 94.2	& 60.6	& \textbf{54.7}	& 60.4	& \textbf{81.8}	& 71.3	& 49.8	& 55.5	& 65.5	& 43.5	& 63.9 \\
\hline
AMMNet & RGB+D & {92.4}	& 88.4	& 82.4	& {61.3}	& \textbf{94.7}	& {65.0}	& {38.9}	& {58.1}	& {76.3}	& {73.2}	& {47.3}	& {46.6}	& {62.0}	& {42.6}	& {60.5} \\
AMMNet$_{\textrm{DLabv3}}$ & RGB+D & \textbf{92.8} & 89.0	& 83.3	& 65.5	& 94.6	& 65.8	& 52.0	& \textbf{64.9}	& 81.1	& \textbf{78.2}	& \textbf{57.6}	& \textbf{66.9}	& \textbf{67.4}	& 45.1	& \textbf{67.2} \\
\hline
\end{tabular}
\vspace{-2mm}
\caption{Result comparisons on the test set of NYUCAD~\cite{firman2016NYUCAD}. Results with “$_{\textrm{DLabv3}}$” denote that these results are based on DeepLabv3~\cite{chen2017rethinking} as the backbone network for RGB image feature extraction. Bold numbers represent the best performance.} 
\label{tab:sotaCAD}
\end{center}}
\vspace{-2mm}
\end{table*}

% -------------------------------------
% -------------------------------------
\begin{figure*}[!t]
\centering
\includegraphics[width=0.99\textwidth]{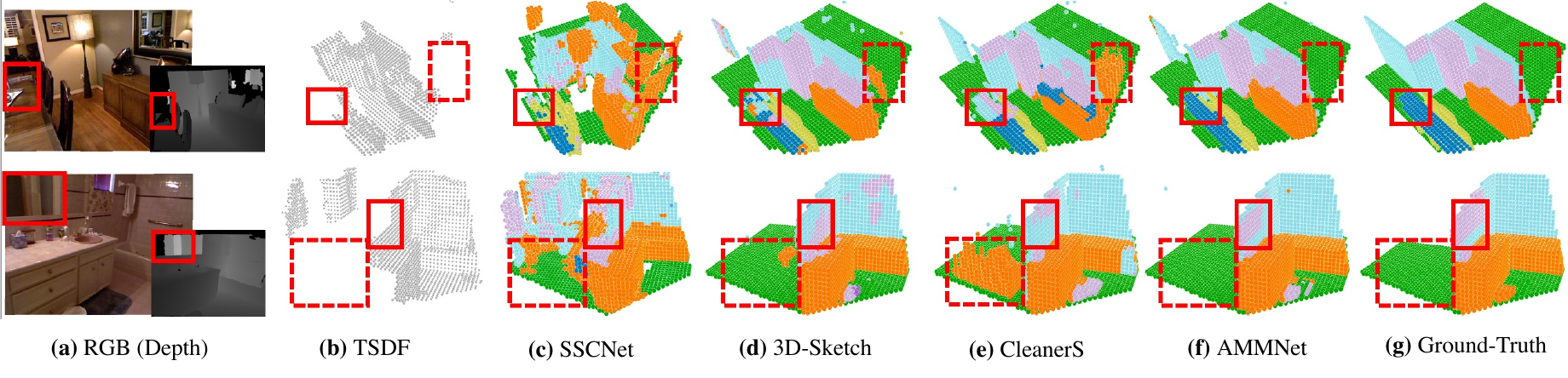}
\vspace{-2mm}
\caption{Qualitative comparison on challenging indoor scenes from the test set of NYU~\cite{silberman2012indoor} with state-of-the-art methods, including SSCNet~\cite{song2017semantic}, 3D-Sketch~\cite{chen20203d}, and CleanerS~\cite{wang2023semantic}. }
\vspace{-3mm}
\label{fig:sota}
\end{figure*}

% -------------------------------------

\section{Experiments}
\label{sec:experiments}

\subsection{Datasets and Evaluation Metrics}

\noindent
\textbf{Datasets.}
Following~\cite{song2017semantic,Garbade2018_twoStream,chen20203d}, we conduct experiments on NYU~\cite{silberman2012indoor} and NYUCAD~\cite{firman2016NYUCAD} datasets which are common indoor benchmarks of SSC. NYU contains 1,449 noisy RGB-D images captured by Kinect sensors. NYUCAD is synthesized from CAD models and has noise-free depth. We use the standard split with 795 images for training and 654 for testing. To select values of hyperparameters, we randomly split one more small set of 100 samples from the training set and take it as a validation set.

\noindent\textbf{Evaluation Metrics.}
Two metrics are used for evaluation: 1) Scene Completion (SC) IoU measuring voxel occupancy prediction accuracy for occluded voxels. It evaluates scene completion capability, and for it, we report precision, recall, and IoU; 2) Semantic Scene Completion (SSC) mIoU measures the accuracy of predicting semantic labels for occupied voxels, where the per-class IoU, as well as the mean IoU (mIoU) across all classes, are reported.

\subsection{Implementation Details}

\noindent
\textbf{Network Architectures.}
For the image encoder, the default setting follows~\cite{wang2023semantic} by adopting Segformer-B2~\cite{xie2021segformer} backbone with four MLP layers for feature extraction. The image encoder initializes from ImageNet~\cite{deng2009imagenet} pre-trained weights and freezes the backbone while keeping the MLP head trainable. 
To facilitate a fair comparison with CVSformer~\cite{dong2023cvsformer}, we alternatively incorporate the pre-trained DeepLabv3 model as the image encoder, which was obtained by training for 1,000 epochs on the RGB image segmentation task and freeze its parameters. The rest of the network trains from scratch.
The TSDF encoder consists of three layers of 3D convolution, two DDR~\cite{li2019rgbd} blocks, and two layers of 3D deconvolution. 
The discriminator consists of four DDR~\cite{li2019rgbd} layers with stride $2, 2, 3, 1$ respectively for downsampling, followed by flattening and two linear layers for prediction. 
The feature channel size $D=256$ and the 3D resolution $G_x,G_y,G_z$ of prediction is set to $(60,36,60)$. 
In the loss function, $\lambda$ is set as 0.25, following ~\cite{wang2023semantic}, $\beta$ is set as 0.005. In the GT perturbations, the probability $P^G, P_j^S$ are set as random variables uniformly distributed between 0.1 and 0.9, a detailed analysis for the hyperparameters is given in Sec.~\ref{ssec:discussion}. 

\noindent
\textbf{Training Settings.}
We implement all experiments using PyTorch on 2 NVIDIA 3090 GPUs. The model is optimized with AdamW~\cite{loshchilov2017decoupled} using a weight decay of 0.05 and an initial learning rate of 0.001. The learning rate is scheduled in a cosine decay policy~\cite{loshchilov2016sgdr} with a minimum value of 1e-7.
Training is conducted for 150 epochs with a batch size of 4. Common data augmentation is applied, including resize, random crop, flip for 2D images, and random x/z axis flip as well as x-z permutation for 3D volumes, following \cite{dourado2022data}.

\subsection{Comparisons with State-of-the-Arts.}
\label{ssec:comp_sota}
To evaluate AMMNet, we compare it against several recent state-of-the-art methods, including single-modal methods SSCNet~\cite{song2017semantic} and CCPNet~\cite{zhang2019cascaded-ccpnet}, and multi-modal methods--DDRNet~\cite{li2019rgbd}, AIC-Net~\cite{li2020anisotropic}, 3D-Sketch~\cite{chen20203d}, FFNet~\cite{wang2022ffnet}, CleanerS~\cite{wang2023semantic}, PCANet~\cite{li2023front}, SISNet~\cite{cai2021semantic}, and CVSformer~\cite{dong2023cvsformer}.

\noindent
\textbf{Quantitative Comparisons.}
1)~\textbf{Results on NYU~\cite{silberman2012indoor}}.
Table \ref{tab:sota} shows AMMNet achieves superior performance on real NYU~\cite{silberman2012indoor} data, outperforming CVSformer~\cite{dong2023cvsformer}, the previous best method, by 3.5\% for SSC-mIoU. Moreover, CleanerS~\cite{wang2023semantic} has the most relevant setting as our AMMNet, using the same RGB and TSDF inputs, and the same baseline architecture, while relying on clean CAD data and a complex distillation pipeline. In contrast, our approach simply unleashes the modality potentials, achieving even better results in a simpler manner.  
2)~\textbf{Results on NYUCAD~\cite{firman2016NYUCAD}}.
Table \ref{tab:sotaCAD} presents results on synthetic NYUCAD~\cite{firman2016NYUCAD} dataset. AMMNet demonstrates noticeable advantages, achieving a top SSC-mIoU of 67.2\%.
Clean CAD environments enable more effective feature learning, thus the benefits of cross-modal modulation become more noticeable, as it can fully exploit the representation power of RGB and TSDF without interference from noise. 

% -------------------------------------
\begin{table}[t]
\footnotesize
\renewcommand\arraystretch{1.3} 
\setlength{\tabcolsep}{6.5pt}{ 
\begin{center}
\begin{tabular}{r| c c c | c c} 
\hline
Methods  & $\mathcal{M}$ & \textbf{$\mathcal{L}_{(D,G)}$} & \textbf{$\mathcal{L}_{SSC}$}   & {{SC-IoU}} & {{SSC-mIoU}} \\
\hline 
Baseline  &  &  & \cmark & 71.6\%  & 46.1\%  \\
\hline
AMMNet  & \cmark &  & \cmark  &  74.0\%  	& 47.4\%   \\
AMMNet  &  & \cmark  & \cmark  & 74.4\%  	& 47.7\%  \\
AMMNet  & \cmark  & \cmark & \cmark  & \textbf{75.6\%}  	& \textbf{48.5\%}  \\
\hline
\end{tabular}
\end{center}}
\vspace{-3mm}
\caption{Ablation studies of different components in AMMNet on the test set of NYU~\cite{silberman2012indoor}.}
\vspace{-2mm}
\label{tab:ablation}
\end{table}

% -------------------------------------
% -------------------------------------
\begin{table}[t]
\footnotesize
\renewcommand\arraystretch{1.3} 
\setlength{\tabcolsep}{3pt}{ 
\begin{center}
\begin{tabular}{r| c | c c c | c c} 
\hline
Methods  & \textbf{$\mathcal{L}_{(D,G)}$} & $\mathcal{M}^{1st}$ & $\mathcal{M}^{2nd}$ & $\mathcal{M}^{3rd}$  & {{SC-IoU}} & {{SSC-mIoU}} \\
\hline
\multirow{3}{*}{AMMNet} & \multirow{3}{*}{\xmark} & \cmark &  &   &  73.5\%  	& 47.1\%   \\
  & & \cmark &         & \cmark    & 74.0\%  	& 47.2\%  \\
 &  & \cmark  & \cmark & \cmark  & 74.0\%  	& 47.4\%  \\
\hline
\multirow{3}{*}{AMMNet} & \multirow{3}{*}{\cmark} & \cmark & &    & 74.7\%  	& 47.8\%  \\
  & & \cmark &         & \cmark     & 74.8\%  	& 48.2\%  \\
 &  & \cmark  & \cmark & \cmark     & \textbf{75.6\%}  	& \textbf{48.5\%}  \\
\hline
\end{tabular}
\end{center}}
\vspace{-3mm}
\caption{Performance comparison of progressively integrating cross-modal modulation modules in AMMNet on the test set of NYU~\cite{silberman2012indoor}, under settings with and without adversarial training. }
\vspace{-2mm}
\label{tab:modulation1}
\end{table}

% -------------------------------------

% \vspace{-2mm}
\noindent
\textbf{Qualitative Comparisons.}
Figure \ref{fig:sota} presents two challenging scene completion examples from the NYU~\cite{silberman2012indoor} dataset to qualitatively compare AMMNet against other methods. AMMNet excels in reconstructing semantically accurate parts, correctly predicting the elongated table and mirror highlighted in red boxes, which other methods fail to capture. Equally importantly, AMMNet completes geometrically plausible structures behind walls, generating empty spaces rather than clutter highlighted in dotted red boxes. Such completion aligns better with natural layouts. Together, these examples showcase AMMNet's strengths - its ability to produce semantically and geometrically faithful scene reconstructions, exceeding prior arts.

\subsection{Ablation Study}
\label{ssec:ablation}

To analyze the contribution of each proposed component, we conduct ablation studies on the NYU~\cite{silberman2012indoor} dataset by adding them incrementally to the baseline model. Following~\cite{wang2023semantic}, the baseline only contains the basic encoder-decoder architecture without cross-modal modulation ($\mathcal{M}$) and adversarial training ($\mathcal{L}_{(D,G)}$). As presented in Table \ref{tab:ablation}, it achieves 46.1\% SSC-mIoU and 71.6\% SC-IoU.

% ------------------------------------
\begin{table}[t]
\footnotesize
\renewcommand\arraystretch{1.3} 
\setlength{\tabcolsep}{6pt}{ 
\begin{center}
\begin{tabular}{r| c c | c c} 
\hline
Methods  & FeatFusion & GradUpdating  & {{SC-IoU}} & {{SSC-mIoU}} \\
\hline
AMMNet$^\star$  & \cmark &     &  73.2\%  	& 46.5\%   \\
AMMNet$^\star$  &  & \cmark    & 74.1\%  	& 46.9\%  \\
AMMNet$^\star$  & \cmark  & \cmark  & \textbf{73.5\%}  	& \textbf{47.1\%}  \\
\hline
\end{tabular}
\end{center}}
\vspace{-3mm}
\caption{Ablation study of two benefits in modulation module $\mathcal{M}^{1st}$ on the test set of NYU~\cite{silberman2012indoor}, under a simplified AMMNet$^\star$ setting where the latter two modulation modules are removed.}
\vspace{-2mm}
\label{tab:modulation2}
\end{table}

% ------------------------------------

\noindent
\textbf{Effectiveness of $\mathcal{M}$.}
When integrating the proposed $\mathcal{M}$ into the baseline model, the performance is improved by 2.4\% in SSC-mIoU and 1.3\% in SC-IoU, as presented in the 2nd row of Table \ref{tab:ablation}. 
Meanwhile, regarding the adversarial trained model in the 3rd row as a strong baseline, integrating $\mathcal{M}$ can still bring an improvement of 1.2\% SC-IoU and 0.8\% SSC-mIoU. This demonstrates the importance of $\mathcal{M}$ for enabling superior semantic scene understanding even over a strong baseline.
% 

% \vspace{-2mm}
\noindent\textbf{Effectiveness of $\mathcal{L}_{(D,G)}$.}
Similarly, adding adversarial training to the baseline improves the SSC-mIoU by 1.6\% and SC-IoU by 2.8\% as shown in the 3rd row of Table \ref{tab:ablation}. 
Without adversarial training in our AMMNet, we observe a performance drop of 1.1\% in SSC-mIoU and 1.6\% in SC-IoU as in the 2nd row. This proves the regularization effect of adversarial training in preventing overfitting.
Furthermore, as depicted in Figure \ref{fig:problem}(b), the performance curves over training epochs validate that incorporating adversarial training leads to steadily increasing SSC-mIoU and SC-IoU on both train and test sets. This demonstrates that the performance gains of adversarial training stem from enabling stable optimization with dynamic competition in gradient updating, instead of potential overfitting to the training data.

% ------------------------------------
\begin{table}[t]
\footnotesize
\renewcommand\arraystretch{1.3} 
\setlength{\tabcolsep}{3.0pt}{ 
\begin{center}
\begin{tabular}{r| c c c | c c} 
\hline
Methods  & Geometry & Semantic  & Discriminator & {{SC-IoU}} & {{SSC-mIoU}} \\
\hline
AMMNet$^\dagger$  &  &    & \cmark &  74.0\%  	& 46.7\%   \\
\hline
AMMNet$^\dagger$  & \cmark &  & \cmark   &  \textbf{74.4}\%  	& 46.6\%   \\
AMMNet$^\dagger$  &  & \cmark  & \cmark  & 74.0\%  	& 47.1\%  \\
AMMNet$^\dagger$  & \cmark  & \cmark & \cmark & \textbf{74.4\%}  	& \textbf{47.7\%}  \\
\hline
\end{tabular}
\end{center}}
\vspace{-3mm}
\caption{Ablation study of the GT perturbations in adversarial training on the test set of NYU~\cite{silberman2012indoor}, under a simplified AMMNet$^\dagger$ setting where all the modulation modules are removed.}
\vspace{-2mm}
\label{tab:gan}
\end{table}

% ------------------------------------

In summary, the ablation studies clearly validate the efficacy of the key components of AMMNet. The combination of both leads to the best performance with an overall improvement of 2.4\% in SSC-mIoU and 4.0\% in SC-IoU.

% ------------------------------------
\CheckRmv{
\begin{figure*}[!t]
\centering
\includegraphics[width=0.99\textwidth]{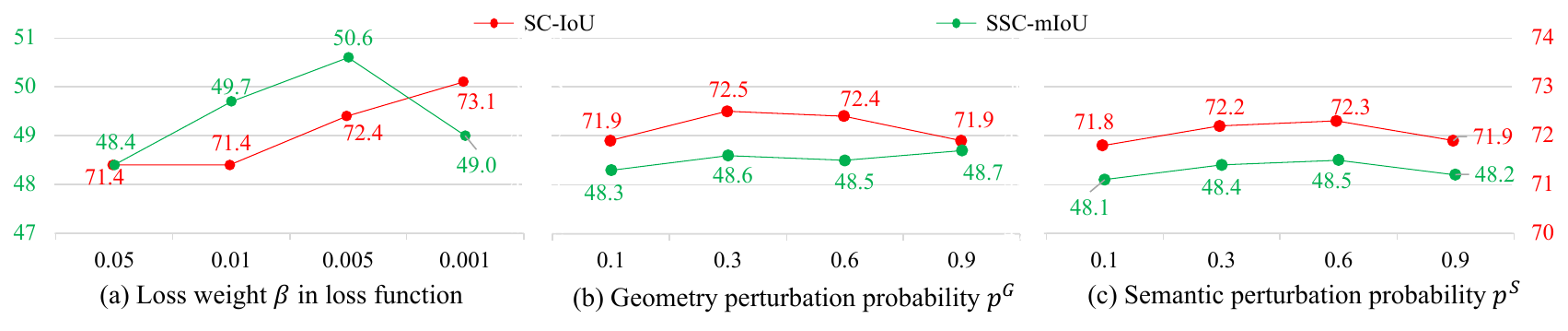}
\vspace{-2mm}
\caption{Sensitivity analysis of key hyperparameters based on the random split validation set from NYU~\cite{silberman2012indoor}.} 
\label{fig:hyper}
\end{figure*}
}

% ------------------------------------

\subsection{Discussions}
\label{ssec:discussion}

\noindent
\textbf{Analysis of Progressive Modulation.}
To further analyze the impact of cross-modal modulation, we conduct experiments by progressively adding modulation components. For simplicity, we denote the three modulation modules in Figure \ref{fig:framework} from left to right as $\mathcal{M}^{1st}$, $\mathcal{M}^{2nd}$ and $\mathcal{M}^{3rd}$ respectively.
As shown in Table \ref{tab:modulation1}, compared with the baseline model, incorporating $\mathcal{M}^{1st}$ brings substantial gains of 1.0\% SSC-mIoU and 1.9\% SC-IoU, validating the benefit of enriching uni-modal features with cross-modal contexts. However, further adding $\mathcal{M}^{2nd}$ and $\mathcal{M}^{3rd}$ only leads to minor improvements over using $\mathcal{M}^{1st}$ alone.
We hypothesize this is because the model capacity is still limited by overfitting. With the proposed adversarial training scheme, progressively integrating more modulation modules consistently improves the performance, as presented in the lower part of Table \ref{tab:modulation1}.

In summary, with proper regularization, the combination of $\mathcal{M}^{1st}$, $\mathcal{M}^{2nd}$ and $\mathcal{M}^{3rd}$ leads to the best results by sufficiently modulating both shallow and deep representations.

% \vspace{-2mm}
\noindent\textbf{Analysis of Modulation Benefits.}
As discussed in Sec.~\ref{sec:method}, compared to addition, modulation provides two benefits: 1) inter-dependent gradient updating across modalities, and 2) enhanced forward feature fusion.
To analyze their contribution, we simplify AMMNet to only have the first modulation $\mathcal{M}^{1st}$, denoted as AMMNet$^\star$.
Based on AMMNet$^\star$, we compare two variants:
1) replacing $\mathcal{M}^{1st}$ with addition but keeping inter-dependent gradient updating;
2) detaching $\mathcal{M}^{1st}$ gradient while retaining forward feature calibration.
As shown in Table \ref{tab:modulation2}, detaching inter-dependent gradient causes an SSC-mIoU drop of 0.6\%, while replacing it with addition leads to a smaller 0.2\% decrease.
This suggests inter-dependent gradient updating provides more contribution, while forward calibration also complements the improvements.

\noindent\textbf{Analysis of Adversarial Training.} 
To thoroughly analyze the gains from adversarial training, we built an ablated AMMNet, denoted as AMMNet$^\dagger$, by removing all cross-modal modulation modules. As shown in Table~\ref{tab:gan}, compared to the baseline model, utilizing the developed discriminator alone (1st row) significantly improves geometric completeness, increasing SC-IoU by 2.4\%. Adding fake samples that disrupt geometric completeness (2nd row) further enhances geometric perception, achieving an additional 0.4\% SC-IoU gain. However, without guidance on semantic validity, it achieves slightly inferior semantic performance compared to the discriminator-alone setting, with a minor 0.1\% SSC-mIoU drop. Introducing fake samples damaging semantic correctness (4th row) provides better semantic supervision, leading to a noticeable improvement in semantic accuracy, increasing SSC-mIoU by 1.1\% over the geometry-only setting. This verifies the complementary effects of the two perturbation strategies in improving scene completion.

\noindent\textbf{Analysis of Hyperparameter Sensitivity.} 
We analyze the sensitivity of the key hyperparameters based on the randomly split validation set: the loss weight $\beta$ in the loss function, and the probabilities $p^G$ and $p^S$ of generating geometrically/semantically perturbed fake GT, where the subscript $j$ in $p_j^S$ is omitted since the same probability is used for perturbing each semantic category here. As Figure~\ref{fig:hyper} (a) shows, an overly large loss weight for the adversarial training loss like 0.05 overwhelms the task-specific loss $\mathcal{L}_{SSC}$, harming the performance. AMMNet selects $\beta=0.005$ which achieves optimal results.

In the sensitivity analysis for the perturbation probabilities $p^G$ and $p^S$, we build on the ablated AMMNet$^\dagger$ with cross-modal modules removed. The experiments show performance remains stable across a wide range of fixed values for both $p^G$ and $p^S$, with only minor variations observed. This indicates the robustness of the proposed adversarial training scheme across different perturbation levels. To avoid hyperparameter tuning burden, we randomly sample probabilities within $[0.1, 0.9]$ for both $p^G$ and $p^S$, avoiding too little noise below 0.1 or too much above 0.9.

\section{Conclusions}
In this work, we identify two limitations of existing RGB-D based semantic scene completion methods: ineffective feature learning and overfitting problems. To address these issues, we propose a new deep learning framework AMMNet. The core techniques include cross-modal modulation to better exploit single-modal representations, and adversarial training to prevent overfitting to the training data. Extensive experiments on NYU and NYUCAD datasets demonstrate AMMNet's state-of-the-art performance.

\section*{Acknowledgements} This research was supported by the National Natural Science Foundation of China under Grant 61925204.
\clearpage
\setcounter{page}{1}
\renewcommand{\thesection}{S\arabic{section}} 
\setcounter{section}{0}
\renewcommand{\thetable}{S\arabic{table}} 
\setcounter{table}{0}
\renewcommand{\thefigure}{S\arabic{figure}}
\setcounter{figure}{0}

\maketitlesupplementary

This supplementary includes ablation results of AMMNet with different image encoders (Sec.~\ref{suppsec_2dnet}), comparison results of using different feature fusion strategies (Sec.~\ref{suppsec_fusion}) and different schemes for alleviating overfitting (Sec.~\ref{suppsec_overfit}), and more visualization results (Sec.~\ref{suppsec_visual}).

\section{Ablation with Different Image Encoder}
\label{suppsec_2dnet}

{\color{red}{This supplementary is for Sec.~\ref{ssec:discussion} of the main paper.}}
To validate the generalization ability of AMMNet, we supplement the ablation study by substituting the SegFormer-B2~\cite{xie2021segformer} image encoder with the ResNet50~\cite{he2016deep} or DeepLabv3~\cite{chen2017rethinking}. As Table \ref{tab:2dnet} shows:
1) adopting the DeepLabv3 encoder pre-trained on image segmentation task achieves the best performance, suggesting that enhanced semantic understanding can better facilitate scene completion and prediction of correct semantic categories. The stronger SegFormer-B2 encoder also improves performance over ResNet50, especially the semantic metric SSC-mIoU by 2.7\% in the baseline model. This aligns with the fact that superior visual features facilitate semantic prediction;
2) AMMNet maintains consistent performance gains over baseline regardless of the choice of image encoder. Specifically, using a ResNet50, AMMNet improves SC-IoU by 3.7\% and SSC-mIoU by 2.3\%. With a SegFormer-B2 encoder, the gains are even higher, reaching 4.0\% and 2.4\% respectively. Notably, AMMNet also achieves considerable improvements of 2.9\% on SC-IoU and 1.5\% on SSC-mIoU with the DeepLabv3 encoder pre-trained on image segmentation task. The consistent gains verify that AMMNet's robust effectiveness stems from a better unleashing of the network potentials, rather than reliance on specific encoders.

% ---------------------------------
\begin{table}[htbp]
\footnotesize
\renewcommand\arraystretch{1.3} 
\newcommand{\tabincell}[2]{\begin{tabular}{@{}#1@{}}#2\end{tabular}}
\setlength{\tabcolsep}{7pt}{ 
\begin{center}
\begin{tabular}{r| c | c | c } 
\hline
Methods & Image Encoder  & {{SC-IoU}} & {{SSC-mIoU}} \\ \hline
Baseline & \multirow{2}{*}{ResNet50}  & 70.3\%  & 43.4\%  \\
AMMNet &  & 74.0\% (\textcolor{red}{$\uparrow$ 3.7\%}) 	& 45.7\% (\textcolor{red}{$\uparrow$ 2.3\%}) \\
\hline
Baseline & \multirow{2}{*}{\tabincell{c}{Segformer\\-B2}}  & 71.6\%  & 46.1\%  \\
AMMNet &  & 75.6\% (\textcolor{red}{$\uparrow$ 4.0\%}) 	& 48.5\% (\textcolor{red}{$\uparrow$ 2.4\%}) \\
\hline
Baseline & \multirow{2}{*}{\tabincell{c}{DeepLabv3}}  & 73.4\%  & 54.6\%  \\
AMMNet &  & 76.3\% (\textcolor{red}{$\uparrow$ 2.9\%}) 	& 56.1\% (\textcolor{red}{$\uparrow$ 1.5\%}) \\
\hline
\end{tabular}
\end{center}}
\vspace{-3mm}
\caption{The ablation study of using different image encoder, including ResNet50~\cite{he2016deep}, Segformer-B2~\cite{xie2021segformer}), and DeepLabv3~\cite{chen2017rethinking}, in our AMMNet on the test set of NYU~\cite{silberman2012indoor}.}
\label{tab:2dnet}
\end{table}
% ---------------------------------

\section{Alternative Fusion Strategies}
\label{suppsec_fusion}

{\color{red}{This supplementary is for Sec.~\ref{ssec:discussion} of the main paper.}}
To validate the efficacy of cross-modal modulation, we compare it with several widely-adopted alternatives for fusing multi-modal representations including addition, concatenation, refinement with SENet~\cite{hu2018squeeze}, refinement with CBAM~\cite{woo2018cbam}, and soft selection~\cite{srivastava2015training}. Experiments are conducted by replacing all three modulation modules in AMMNet with each scheme. Due to the enormous computational overhead of 3D tasks, we do not consider transformer-based attention methods.

As Table~\ref{tab:featfusion} shows, simple fusion schemes like direct addition or concatenation prove insufficient for optimally exploiting cross-modal representations. Incorporating refinement as SENet~\cite{hu2018squeeze} provides a slight 0.3\% SSC-mIoU improvement over the addition. Another alternative, dynamically selecting modalities via soft gating~\cite{srivastava2015training}, provides 0.6\% SSC-mIoU gain over baseline addition. Our proposed cross-modal modulation obtains further noticeable performance gains, elevating SSC-mIoU by 0.6\% and SC-IoU by 0.8\% over incorporating soft selection~\cite{srivastava2015training}. Experiments validate cross-modal modulation facilitates more holistic fusion to better discover synergistic cross-model potential.

% ---------------------------------
\begin{table}[htbp]
\footnotesize
\renewcommand\arraystretch{1.3} 
\setlength{\tabcolsep}{11pt}{ 
\begin{center}
\begin{tabular}{ c c | c c } 
\hline
{Method}  & {Fusion Type} & {{SC-IoU}} & {{SSC-mIoU}} \\
\hline
- &  Add & {75.2\%}  	& 47.3\%   \\
-  & Concat  & 75.0\%  	& 46.8\%  \\
\hline
SENet~\cite{hu2018squeeze} & Refine & 74.2\%  	& 47.6\%  \\
CBAM~\cite{woo2018cbam} & Refine & 74.6\% & 47.4\% \\
\hline
HighWay~\cite{srivastava2015training}  & SoftSelect   & 74.8\% & 47.9\% \\
Ours & Modulation & \textbf{75.6\%} & \textbf{48.5\%} \\
\hline
\end{tabular}
\end{center}}
\vspace{-3mm}
\caption{Performance comparison of different feature fusion schemes by replacing all three modulation modules in AMMNet with each scheme respectively.}
\label{tab:featfusion}
\end{table}

% ---------------------------------

% ---------------------------------
\begin{table}[b!]
\footnotesize
\renewcommand\arraystretch{1.3} 
\setlength{\tabcolsep}{5pt}{ 
\begin{center}
\begin{tabular}{ c | c | c c } 
\hline
{Method}  & {Removed Scheme} & {{SC-IoU}} & {{SSC-mIoU}} \\
\hline
AMMNet$^\dagger$ &  None & 74.4\%  	& 47.7\%   \\
\hline
AMMNet$^\dagger$  & Dropout  &  74.7\% ({$\uparrow$ 0.3\%})	& 47.5\% (\textcolor{red}{$\downarrow$ 0.2\%}) \\
AMMNet$^\dagger$ & Label Smooth &  73.4\% (\textcolor{red}{$\downarrow$ 1.0\%})	& 47.3\% (\textcolor{red}{$\downarrow$ 0.4\%}) \\
AMMNet$^\dagger$ & Data Augment(3D) & 74.3\% (\textcolor{red}{$\downarrow$ 0.4\%}) & 47.4\% (\textcolor{red}{$\downarrow$ 0.1\%})\\
AMMNet$^\dagger$ & Data Augment(2D) & 74.8\% (\textcolor{red}{$\uparrow$ 0.1\%}) & 47.0\% (\textcolor{red}{$\downarrow$ 0.7\%}) \\
AMMNet$^\dagger$ & $\mathcal{L}_{(D,G)}$ & 71.6\% (\textcolor{red}{$\downarrow$ 2.8\%}) & 46.1\% (\textcolor{red}{$\downarrow$ 1.6\%}) \\
\hline
\end{tabular}
\end{center}}
\vspace{-3mm}
\caption{The ablation study of different schemes for alleviating overfitting based on AMMNet$^\dagger$ on the test set of NYU~\cite{silberman2012indoor}.}
\label{tab:overfit}
\end{table}

% ---------------------------------

\section{Schemes to Alleviate Overfitting}
\label{suppsec_overfit}

% ---------------------------------
\begin{figure*}[t!]
\centering
\includegraphics[width=0.99\textwidth]{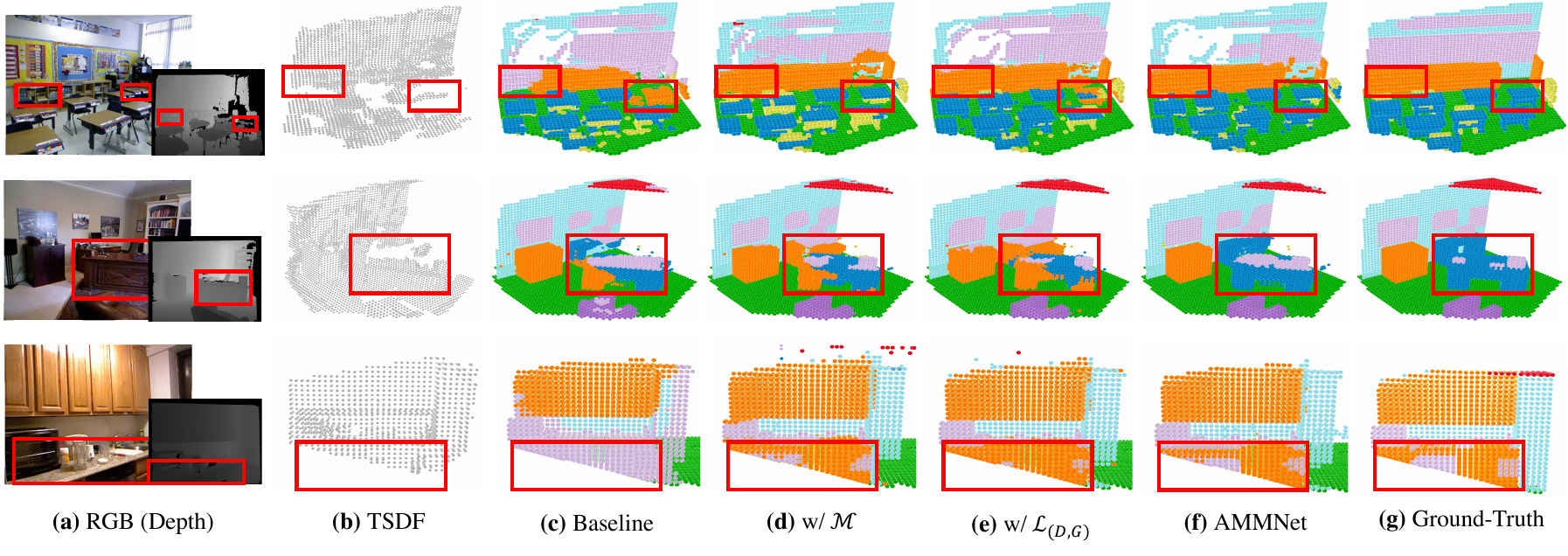}
\vspace{-2mm}
\caption{Visualization results for ablation study based on the test set of NYU~\cite{silberman2012indoor}. The proposed cross-modal modulation $\mathcal{M}$ (in (d)) and adversarial training scheme $\mathcal{L}_{(D,G)}$ (in (e)) improve the baseline with better volumetric occupancy and semantics. Combining both (in (f)) achieves the best results.}
\label{fig:ablation}
\end{figure*}

% ---------------------------------

% ---------------------------------
\begin{figure*}[t!]
\centering
\includegraphics[width=0.99\textwidth]{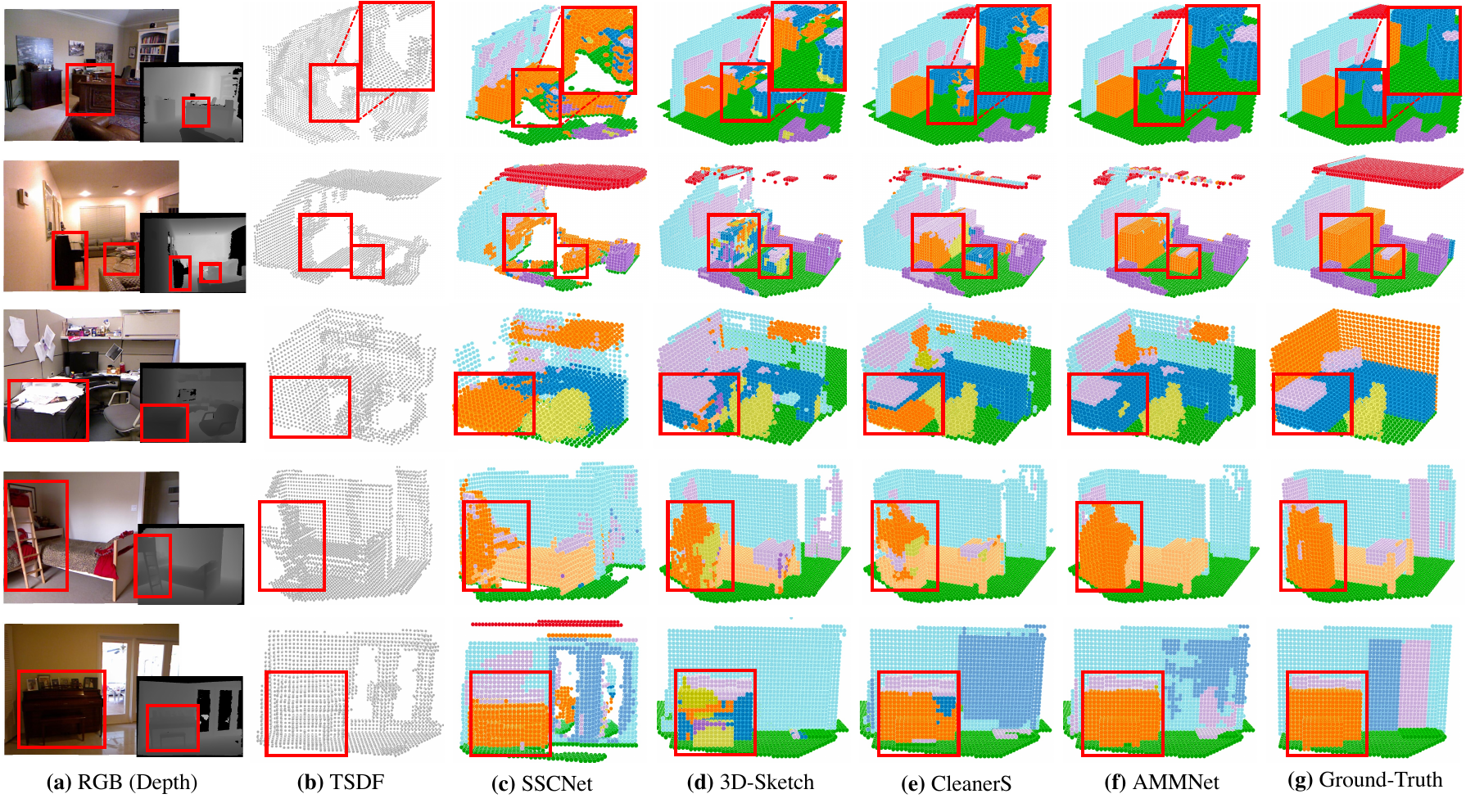}
\vspace{-2mm}
\caption{More qualitative comparisons on challenging indoor scenes from the test set of NYU~\cite{silberman2012indoor} with state-of-the-art methods, including SSCNet~\cite{song2017semantic}, 3D-Sketch~\cite{chen20203d}, and CleanerS~\cite{wang2023semantic}. }
\label{fig:sota2}
\end{figure*}

% ---------------------------------

{\color{red}{This supplementary is for Sec.~\ref{ssec:discussion} of the main paper.}} 
To examine the isolated impact of different schemes in alleviating overfitting, we ablate different modules from the AMMNet$^\dagger$, where all cross-modal modulation modules are removed. Schemes analyzed include dropout, label smoothing, 2D/3D data augmentation, and our adversarial training scheme $\mathcal{L}_{(D,G)}$. As Table~\ref{tab:overfit} reports, removing most regularizers causes minor performance drops, confirming their auxiliary effects. Specifically, simple schemes like dropout exhibit weaker regularization power, as evaluated by the minor 0.2\% SSC-mIoU drop when excluded. Meanwhile, complementary strategies like label smoothing~\cite{szegedy2016rethinking} (0.4\% SSC-mIoU drop) and 2D/3D augmentation (0.1\%/0.7\% SSC-mIoU reduction) help prevent learned biases and memorization. Findings confirm the necessity of strong regularization guided by domain insights. 

Notably, excluding our $\mathcal{L}_{(D,G)}$ degrades results substantially by 2.8\% in SC-IoU and 1.6\% SSC-mIoU. This verifies the vital role of our custom adversarial training scheme in alleviating overfitting. We advise blending it with existing methods like augmentation and label smoothing~\cite{szegedy2016rethinking} to maximize performance.

\section{More Visualization Results}
\label{suppsec_visual}

{\color{red}{This supplementary is for Sec.~\ref{ssec:ablation} and Sec.~\ref{ssec:comp_sota} of the main paper.}}
As Figure~\ref{fig:ablation} shows, incorporating the proposed cross-modal modulation $\mathcal{M}$ (in (d)) improves semantic perception over the baseline (in (c)), correcting erroneous predictions. Building on this, additionally introducing adversarial training (our AMMNet in (f)) further unleashes model potentials, attaining high-fidelity outputs better approximating the ground truth voxels (in (g)). In Figure~\ref{fig:sota2}, we supplement more visual examples compared to state-of-the-art methods.

{
    \small
    \bibliographystyle{ieeenat_fullname}
    \bibliography{main}
}


\end{document}